# Checkerboard artifact free sub-pixel convolution

A note on sub-pixel convolution, resize convolution and convolution resize

Andrew Aitken*, Christian Ledig*, Lucas Theis*, Jose Caballero, Zehan Wang, Wenzhe Shi*
Twitter, Inc.[1]

Convolutional neural networks (CNNs) are a popular and highly performant choice for pixel-wise dense prediction or generation. One of the commonly required components in such CNNs is a way to increase the resolution of the network's input. The lower resolution inputs can be, for example, low-dimensional noise vectors in image generation [7] or low resolution (LR) feature maps for network visualization [4]. Originally described in Zeiler et al. [3], a network layer performing this upscaling task is commonly referred to as a "Deconvolution layer", and has been used in a wide range of applications including super-resolution [1], semantic segmentation [5], flow estimation [6] and generative modeling [7]. The deconvolution layer can be described and implemented in various ways. This led to many names that are often used synonymously, including sub-pixel or fractional convolutional layer [7], transposed convolutional layer [8,9], inverse, up, backward convolutional layer [5,6] and more.

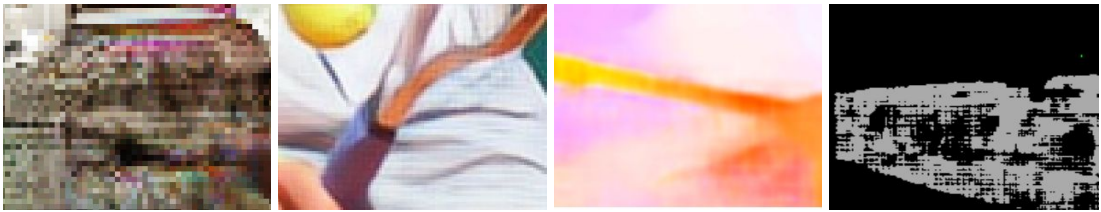

Figure 1: from left to right, varying degrees of checkerboard artifacts in generated images [7], super-resolved images [13], dense flows [6] and dense labels [24]. Figures adapted from respective papers.

The most prominent problem associated with the deconvolution layer is the presence of checkerboard artifacts in output images and dense labels as shown in Figure 1. To combat this problem, smoothness constraints, post processing and different architecture designs have been proposed [6,13,24]. Odena et al. [2] highlight three sources of checkerboard artifacts: *deconvolution overlap*, *random initialization* and *loss functions*.

When the kernel size of the deconvolution is not divisible by the stride (upscaling factor), the number of LR features that contribute to a single high resolution (HR) feature is not constant across the HR feature maps; this is called the *deconvolution overlap*. Odena et al. [2] further observe that even when the kernel size is divisible by the stride, after random initialization the HR output still contains checkerboard artifacts. In this note, we call this effect the "*random initialization*" problem. We further discuss the reason why checkerboard artifacts are present after random initialization in Section 1. Finally, downsampling operations such as strided convolutions or max-pooling in the loss function (e.g. VGG feature loss [23] or the discriminator network in a GAN) can result in inhomogeneous gradient updates.

The first two sources of artifacts, deconvolution overlap and random initialization, can be eliminated by the resize convolution proposed by Odena et al. [2]. Resize convolution first upscales the LR feature maps using nearest-neighbor (NN) interpolation and then employs a standard convolutional layer with

---

both input and output in HR space. Resize convolutions became a popular choice for generative modeling [15, 16] to alleviate checkerboard artifacts.

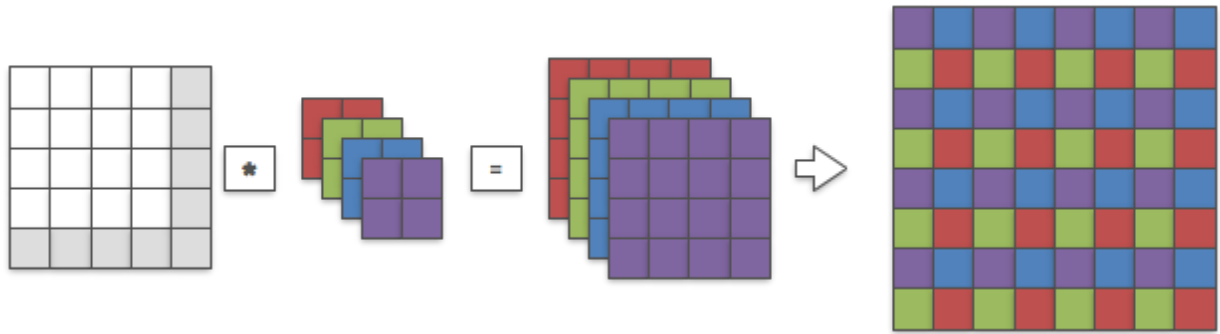

Figure 2: Sub-pixel convolution can be interpreted as convolution + shuffling.

Sub-pixel convolution [1,14] is a specific implementation of a deconvolution layer that can be interpreted as a standard convolution in low-resolution space followed by a periodic shuffling operation as shown in Figure 2. Sub-pixel convolution has the advantage over standard resize convolutions that, at the same computational complexity, it has more parameters and thus better modelling power [14]. Sub-pixel convolution is constrained to not allow deconvolution overlap [2,14], however it suffers from checkerboard artifacts following random initialization as shown in Figure 3.

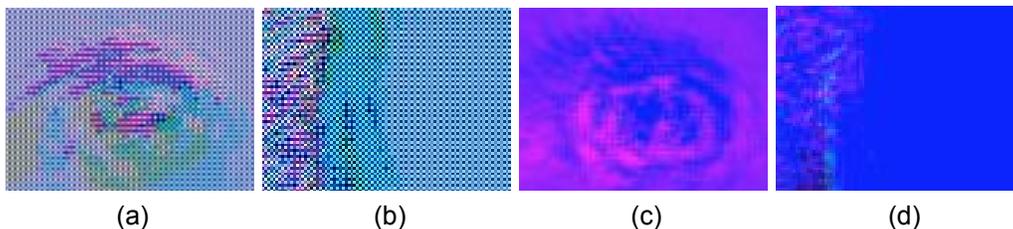

(a)          (b)          (c)          (d)

Figure 3: Two super-resolved images after orthogonal initialization. (a-b) show the results for sub-pixel convolution. (c-d) show the results for NN resize convolution.

In this note we propose a new initialization scheme for sub-pixel convolution, which alleviates the problem of checkerboard artifacts due to random initialization. We demonstrate for image super-resolution that the proposed scheme benefits from the additional modelling power of the sub-pixel convolution compared to the resize convolution, while also removing checkerboard artifacts after initialization.

This note is organized as follows: In Section 1 we introduce notations, recap sub-pixel convolution [14] and resize convolution [2]. In Section 2 we describe the proposed initialization scheme before we present experimental results in Section 3. A brief discussion of our findings is provided in Section 4.

**Section 1: Sub-pixel convolution and resize convolution**

We consider a super-resolution problem in which an original color HR Image, denoted as $I^{HR}$, is downsampled using bicubic resampling by a scale factor of $r$ in each dimension to produce a corresponding LR image, denoted as $I^{LR}$. We aim to train a CNN that generates super-resolved images, denoted as $I^{SR}$, as similar as possible to $I^{HR}$ at the original resolution. In this note, we set $r = 2$ and employ a 5 block ResNet with skip connection [19], similar to the one used in [22]. All convolutional layers except the final layer have 64 channels and 3x3 filter kernels. The final convolution layer performs the following operations:

$$I^{SR} = P(W * f^{L-1}(I^{LR}) + b), \quad (1)$$

Here $P$ is the periodic shuffling operation defined in Shi et al. [1], $W$ is the convolution kernel, $b$ is the bias and $f^{L-1}$ is network's output before the last layer. As $b$ is usually initialized such that all elements are 0, in the following we will neglect $b$ to improve readability. For our particular image SR problem with upscaling factor 2, the kernel $W$ of size (12, 64, 5, 5) has 12 output channels obtained with 5x5 filters. Those 12 output channels are then reorganized by $P$ into $I^{SR}$ with 3 output channels (one for each color). For resize convolution, to match the computation, we resize the activation $f^{L-1}$ with nearest neighbor interpolation and then output 3 channels with 5x5 filters. We used orthogonal initialization [20] as default initialization scheme.

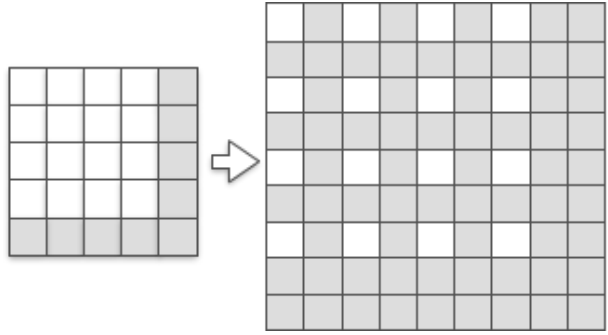

Figure 4: How to create a sub-pixel space from LR space, where grey indicates zeros.

Before introducing our new initialization, we need to take a closer look at the sub-pixel convolution and resize convolution. A detailed description of the relationship between the deconvolution, transposed convolution and sub-pixel convolution can be found in Shi et al. [14]. Similarly, a detailed description of the relationship between the deconvolution and resize convolution can be found in Odena et al. [2]. The sub-pixel convolution, as described in Shi et al. [14], has two different interpretations. The first one is to consider it as a normal convolution followed by a periodic reshuffling, as illustrated in Figure 2 and Equation 1. The second interpretation is to view sub-pixel convolution as a convolution in sub-pixel space. In the sub-pixel image $I^{sp}$, void pixels filled with zeros are created between the original pixels as illustrated in Figure 4. These two interpretations are identical because we can recreate one sub-pixel space convolution kernel $W^{sp}$ from $r^2$ convolution kernels $W$ in LR space [14]. We can then express Equation 1 as:

$$P(W * f^{L-1}(I^{LR})) = W^{sp} * SP(f^{L-1}(I^{LR})), \quad (2)$$

where $SP$ is the operation that transforms LR space into sub-pixel space. In the appendix, we show the sub-pixel space convolution kernel $W^{sp}$ with size (3, 64, 10, 10) that is obtained from the last layer's convolution kernel $W$ with size (12, 64, 5, 5) straight after the orthogonal initialization.

To allow for a more concise description of the proposed initialization scheme, we now define the term "sub kernel". For zero-indexed matrices we consider a convolution kernel $W$ with size $(c_o, c_i, w, h)$ where $c_o$ is an integral multiple of the squared rescale factor $r^2$. A sub kernel is one 2D convolution kernel $W(o, i, :, :)$ where $o$ is the output and $i$ the input channel.

For a given $k \in \{0, ..., c_o/r^2 - 1\}$ a group of $r^2$ consecutive sub kernels, $W_k = \{W(kr^2 + n, i, :, :) : n \in \{0, ..., r^2 - 1\}\}$ of size $(r^2, 1, w, h)$ recreates one sub-pixel space convolution kernel $W^{sp}(k, i, :, :)$ with size $(1, 1, wr, hr)$. More importantly, we can also define for a given $n \in \{0, ..., r^2 - 1\}$ a group of sub kernels $W_n = W(kr^2 + n, :, :, :) : k \in \{0, ..., c_o/r^2 - 1\}\}$. We now employ this definition of the sub kernel sets $W_n$ to explain the appearance of checkerboard artifacts after model initialization. Sub-pixel convolution can be viewed as:

$$I^{SR}{}_n = W_n * f^{L-1}(I^{LR}) , \quad (3)$$

followed by $P$ which rearranges all $I^{SR}{}_n$ to be $I^{SR}$. This means that each of the $r^2$ sub kernel sets $W_n$ is responsible for only one subset of $I^{SR}$. Conversely, each pixel in $I^{SR}$ only depends on one sub kernel set $W_n$. The fact that each sub kernel set $W_n$ is initialized independently but applied to the same input $f^{L-1}(I^{LR})$ to generate neighboring features in HR space causes the checkerboard artifacts in the generated images after initialization. An example of this problem is shown in Figure 3.

Nearest neighbor (NN) resize convolution can be interpreted as filling the sub-pixel space with nearest neighbor interpolation instead of zero. This is then followed by a convolution in sub-pixel space to produce HR outputs:

$$I^{SR} = W^{sp} * N(f^{L-1}(I^{LR})) , \quad (4)$$

where $N$ is the NN resize operation. It is clear that this approach resolves the problem of deconvolution overlap since the stride is always 1. It is also clear that effects caused by random initialization are removed since all kernel weights are activated for each calculated HR feature as shown in Figure 3. However, convolution with zero is skipped in sub-pixel convolution but convolution with NN interpolated pixels can not be skipped in resize convolution. As a result, for the same complexity, the approach based on sub-pixel convolution has more trainable parameters [14].

In the following we describe an initialization scheme that does not rely on convolutions in high resolution space while still being unaffected by random initialization. This allows us to benefit from the efficiency of sub-pixel convolutions while avoiding checkerboard patterns caused by initialization.

**Section 2: Initialization to convolution resize**

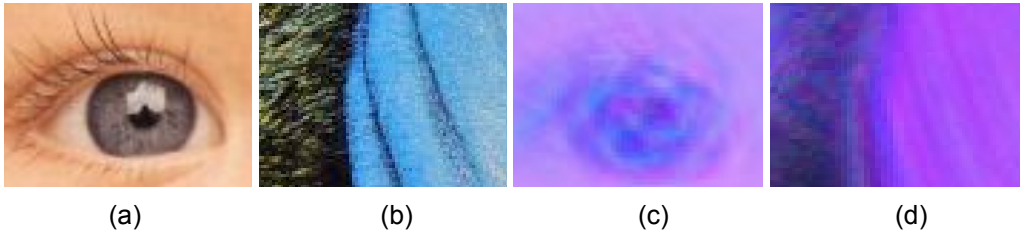

(a)　　　　　　　(b)　　　　　　　(c)　　　　　　　(d)

Figure 5: Two super resolved images after repeat sub kernel initialization. (a-b) show the original HR images. (c-d) show the results for repeat sub kernel initialized sub-pixel convolution.

One alternative to NN resize convolution, in terms of eliminating checkerboard artifacts, is to swap the order of the operations and to first perform the convolution which is then followed by nearest neighbor resizing. We can then express the reconstructed image $I^{SR}$ as the output of a CNN followed by NN resize:

$$I^{SR} = N(W * f^{L-1}(I^{LR})), \quad (5)$$

In contrast to the reshuffling operator $P$, which reduces the number of feature maps by a factor of $r^2$, NN resize preserves the number of feature maps while increasing their spatial resolution. Instead of considering $r^2$ sub kernel sets $W_n$ we only need a single set $W_0$ and are thus no longer able to learn the upscaling operation. Using the notation that we introduce in Section 1 we can then write $I^{SR}$ as:

$$I^{SR} = N(W_0 * f^{L-1}(I^{LR})) : \forall n, \ I^{SR}_n = W_0 * f^{L-1}(I^{LR}). \quad (6)$$

While this approach guarantees checkerboard free reconstructions after initialization it has obvious drawbacks in that the upsampling kernel is not trainable, unlike the sub-pixel and resize convolutions. The important insight is, however, that to eliminate checkerboard patterns, we only need to ensure that the sub-pixel convolution is identical to convolution NN resize after initialization. This means that we aim to find initial weights $W'$ so that the following is true after initialization:

$$P(W' * f^{L-1}(I^{LR})) = N(W_0 * f^{L-1}(I^{LR})). \quad (7)$$

Due to Equation 6 and the following relation for general kernels W':

$$I^{SR} = P(W' * f^{L-1}(I^{LR})) : \forall n, \ I^{SR}_n = W'_n * f^{L-1}(I^{LR}), \quad (8)$$

we can prevent checkerboard artifacts after initialization by setting $\forall n : W'_n = W_0$. In practice, this means that we only need to initialize $W_0$ and then copy the weights to the rest of the sub kernel sets $W'_n$. This effectively means that $W^{sp'} = N(W_0)$. We show some examples of the initialized $W^{sp'}$ in the Appendix and demonstrate that the network is free from checkerboard artifacts immediately after initialization in Figure 5.

**Section 3: Experiments**

To demonstrate the effectiveness of the proposed initialisation scheme, we trained several super-resolution models for the previously described super-resolution task. All networks were trained using a random sample of 350 thousand images from the ImageNet database [17]. The input images were in BGR format normalized to 0 to 1, and the output images in BGR format normalized to -1 to 1. During training for each mini-batch we crop 16 random 96x96 HR sub-images from different training images. For optimization we use Adam [12] with $\beta_1 = 0.9$. The networks were trained with a learning rate of $10^{-4}$ for $8 * 10^6$ update iterations. The training time on a single M40 GPU is approximately 7 days. Mean squared error (MSE) between $I^{HR}$ and $I^{SR}$ was used as the loss function to assess both training and testing error. The training losses were saved every iteration and plotted as an average across 350K crops, equivalent to one epoch. The test errors were saved every $10^5$ iterations and calculated as an average across 200 images from the test set of BSD500 [18].

Training losses and testing errors from the sub-pixel convolution, NN resize convolution and sub-pixel convolution initialized to convolution NN resize (ICNR) are shown in Figure 7 and Figure 8. This supports our hypothesis that the extra parameters of the sub-pixel convolution allow it to outperform the NN resize convolution substantially. Interestingly, sub-pixel convolution initialized to convolution NN converges faster and to a better minimum as compared to standard initialization. This result implies that the proposed initialization, which removes checkerboard artifacts, provides a more reasonable starting point for training compared to random initialization.

This is also supported by the visualization of the $W^{sp}$ kernels in the Appendix. It is clear that at the beginning of the training, up until 2 million iterations, kernels initialized with the proposed scheme appear substantially smoother compared to kernels initialized with standard orthogonal initialization. After 8 million iterations, the differences between the kernels appear to reduce and all three sets of kernels converge to similar patterns.

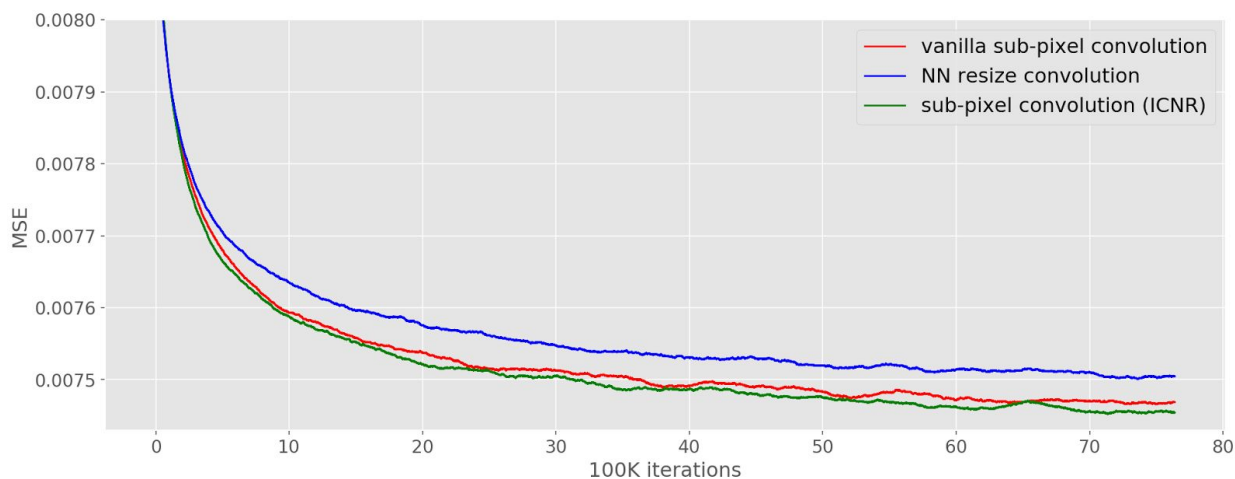

Figure 7: Training loss over time

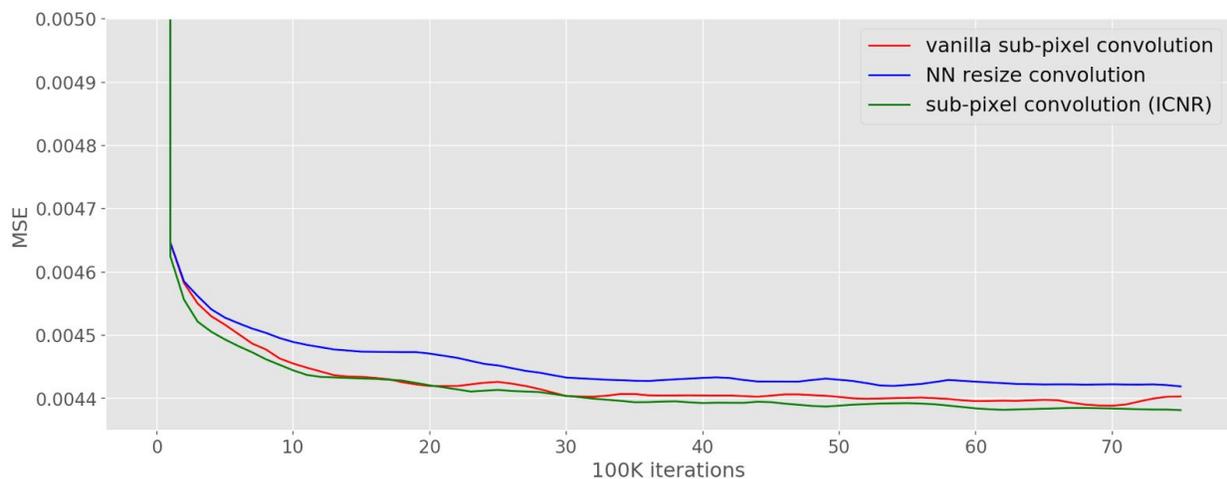

Figure 8: Test loss over time

**Section 4: Discussion**

In this note, we proposed an initialization method for sub-pixel convolution known as convolution NN resize. Compared to sub-pixel convolution initialized with schemes designed for standard convolution kernels, it is free from checkerboard artifacts immediately after initialization. Compared to resize convolution, at the same computational complexity, it has more modelling power and converges to solutions with smaller test errors.

One potential future direction is to replace the NN resize in Equation 7 with a more powerful upscaling operation $F$:

$$P(W' * f^{L-1}(I^{LR})) = F(W_0 * f^{L-1}(I^{LR})) \ . \qquad (9)$$

In most cases such as bicubic and bilinear resize, $F$ can be interpreted as nearest neighbour upsampling followed by convolution with a smooth kernel $B$. With $F(X) = B * N(X)$ and interpreting subpixel convolution as convolution in subpixel-space (c.f. Equation 2), we can write Equation 9 as:

$$W^{sp\prime} * SP(f^{L-1}(I^{LR})) = B * N(W_0 * f^{L-1}(I^{LR})) \ . \qquad (10)$$

With the following equivalence for NN resize:

$$N(W_0 * f^{L-1}(I^{LR})) = N(W_0) * SP(f^{L-1}(I^{LR})) \ , \qquad (11)$$

we can establish a direct relationship between $W^{sp\prime}$ and $B * N(W_0)$ and rewrite Equation 10 as:

$$W^{sp\prime} * SP(f^{L-1}(I^{LR})) = B * N(W_0) * SP(f^{L-1}(I^{LR})) \ . \qquad (12)$$

Because convolution with infinite support is associative, we can fulfill Equation 10 by initializing $W^{sp\prime}$ by $W^{sp\prime} = B * N(W_0)$. However, for non-trivial upscaling kernels $B$ this means that $W^{sp\prime}$ will have larger spatial dimensions compared to initialized to convolution NN resize. While we are still exploring ways to exploit this relation for more efficient initialization approaches, we want to share these first insights with the community.

As discussed by Odena et al. [2], if strided convolution or pooling are present in networks that are used as part of the loss function, then checkerboard artifacts are likely to appear in the generated/predicted images and labels. This additional source of artifacts is not addressed in this work, as all experiments were performed with a simple MSE loss. The solution to this challenging problem remains an open question.

**Appendix:**

In this section, we show the evolution of the $W^{sp}$ kernels from sub-pixel convolution (SPC), resize convolution and SPC initialized to convolution NN resize. The kernel size is $(3, 64, 10, 10)$ for sub-pixel convolution and $(3, 64, 5, 5)$ for resize convolution. Note that 10x10 sub-pixel convolution kernels initialized to convolution NN resize appear due to their initialization constraint like 5x5 kernels. The first 3 channels are converted to BGR for visualization purposes.

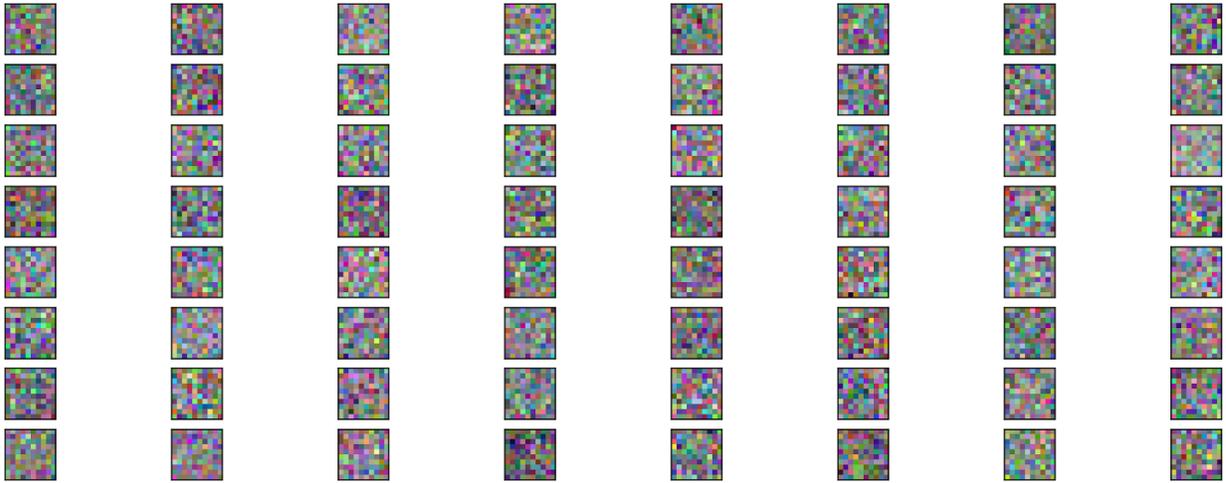

Iteration 0: sub-pixel convolution: $W^{sp}$ with 64 kernels of size 10x10

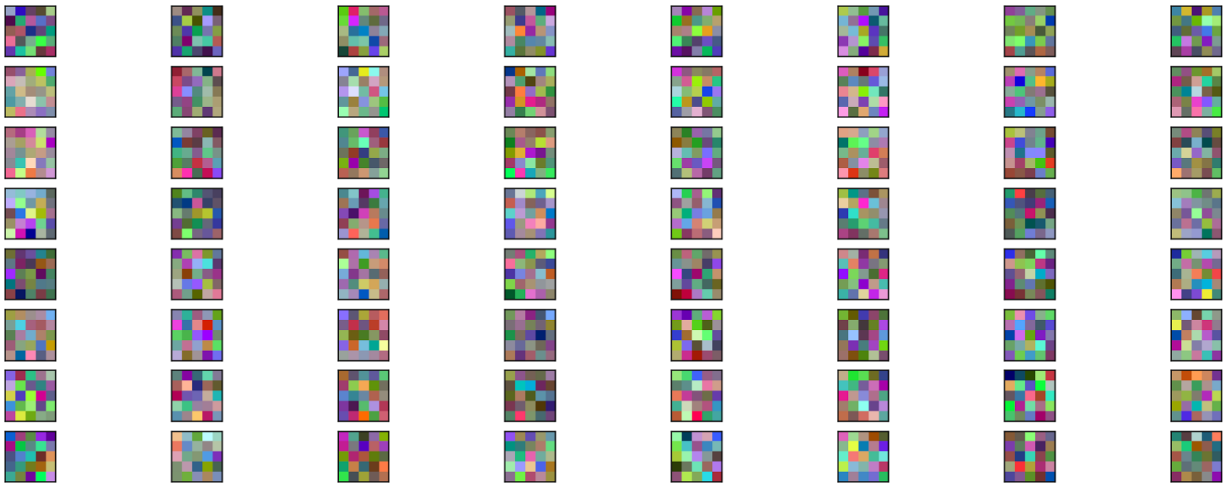

Iteration 0: SPC initialized to convolution NN resize: $W^{sp}$ with 64 kernels of size 10x10

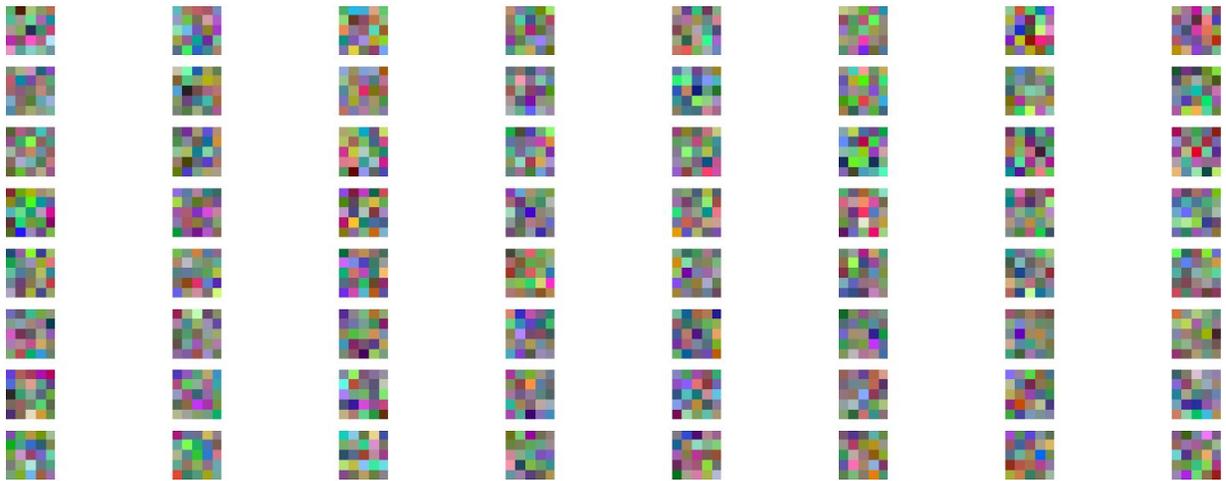

Iteration 0: resize convolution: $W^{sp}$ with 64 kernels of size 5x5

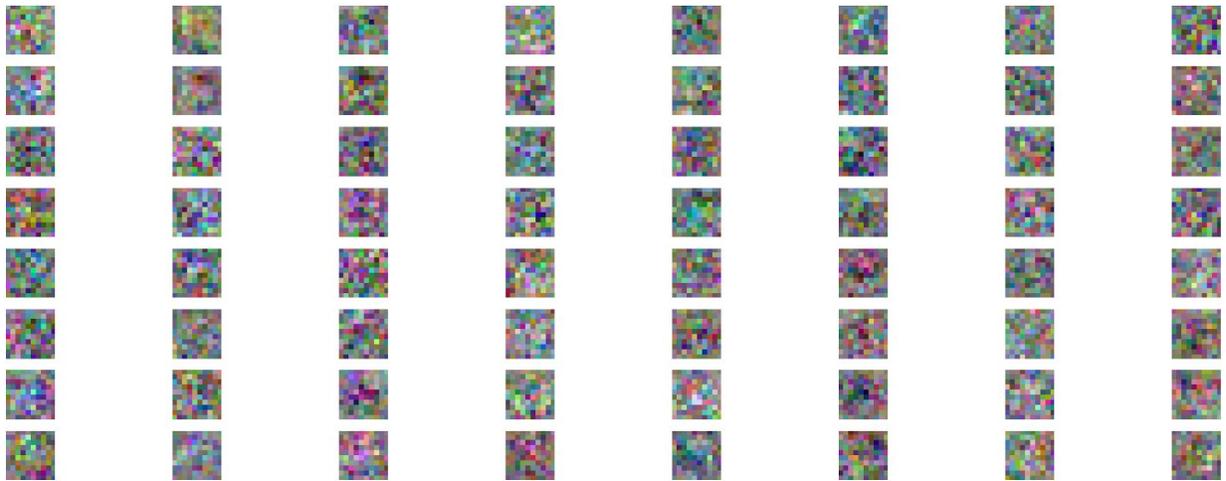

Iteration 100:000, SPC: $W^{sp}$ with 64 kernels of size 10x10

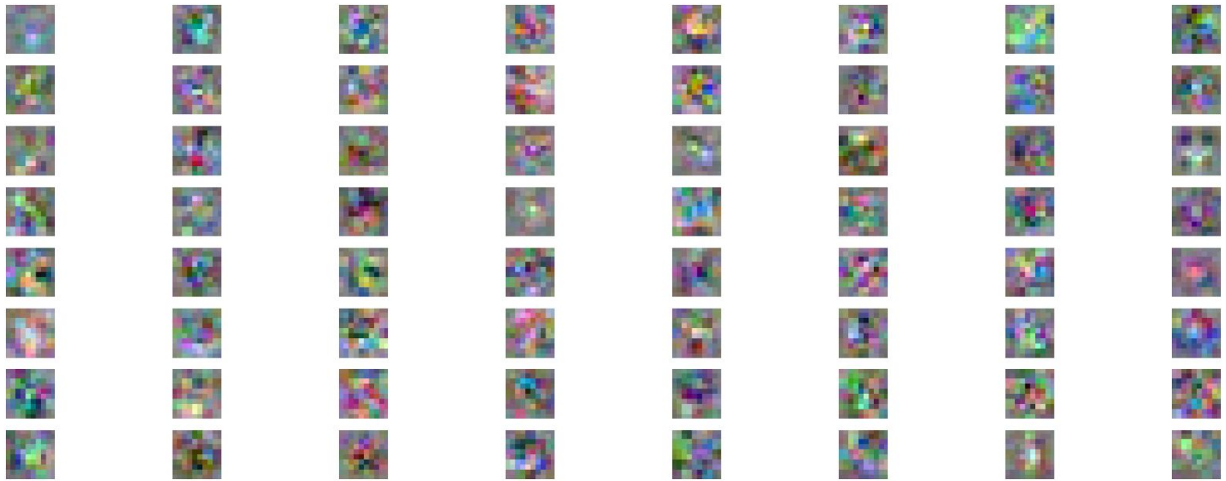

Iteration 100,000, SPC initialized to convolution NN resize: $W^{sp}$ with 64 kernels of size 10x10

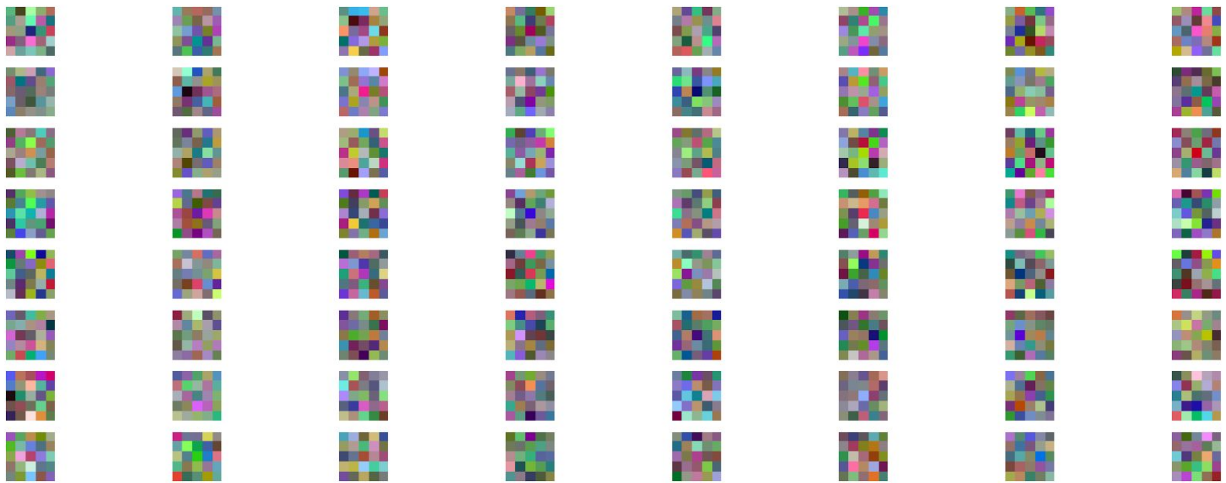

Iteration 100,000, resize convolution: $W^{sp}$ with 64 kernels of size 5x5

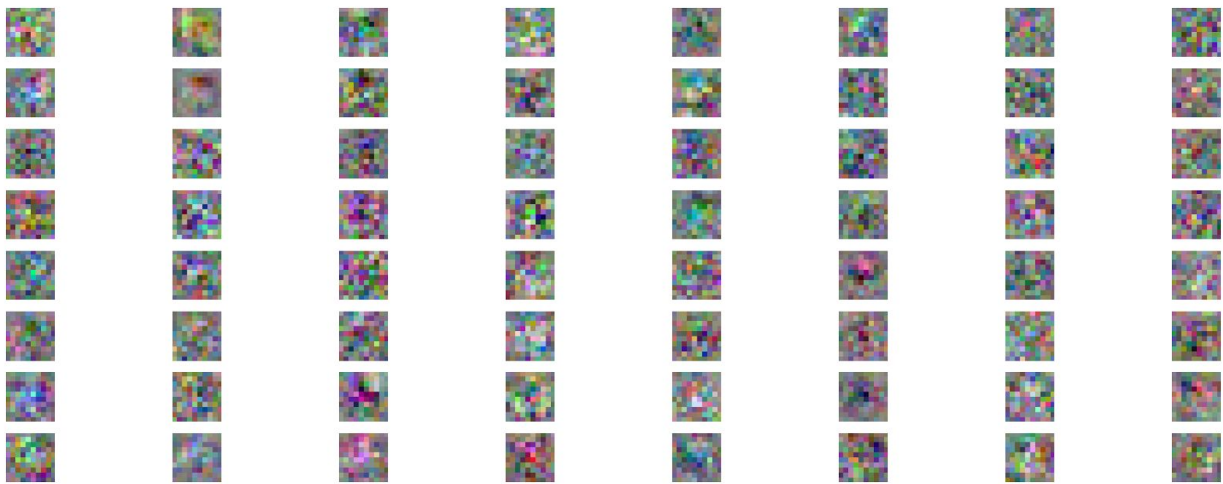

Iteration 200,000, SPC: $W^{sp}$ with 64 kernels of size 10x10

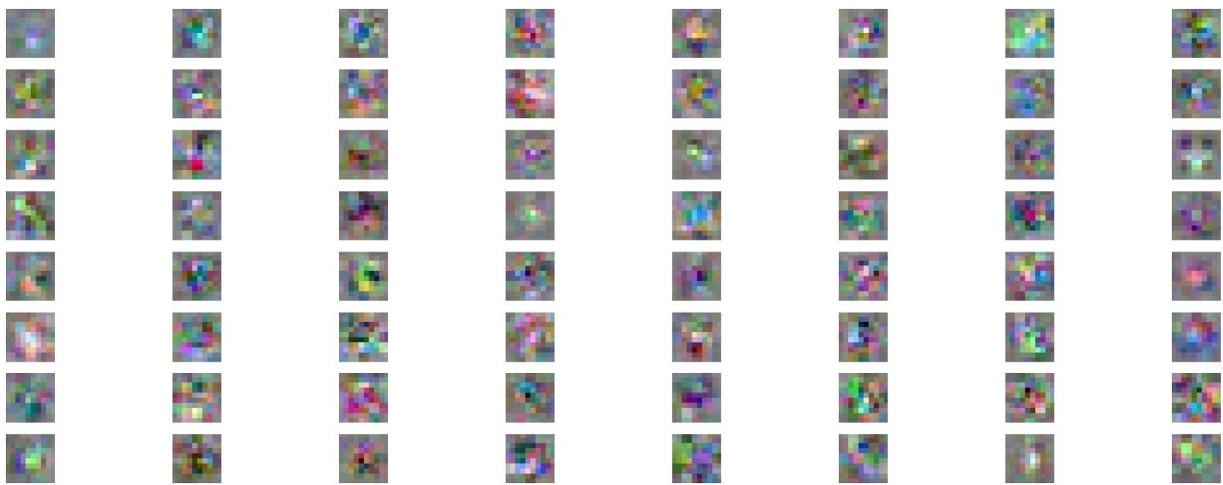

Iteration 200,000, SPC initialized to convolution NN resize: $W^{sp}$ with 64 kernels of size 10x10

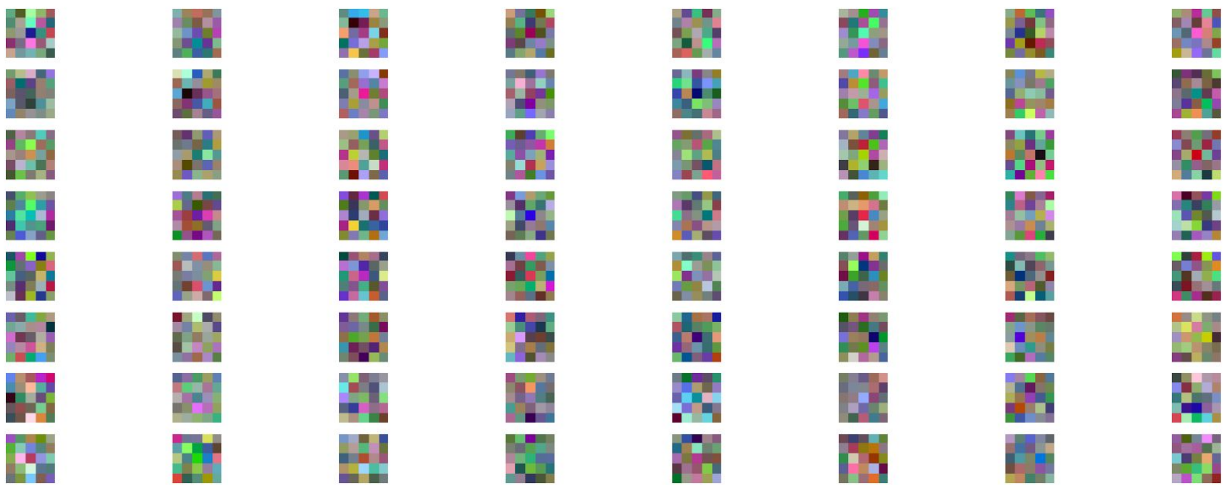

Iteration 200,000, resize convolution: $W^{sp}$ with 64 kernels of size 5x5

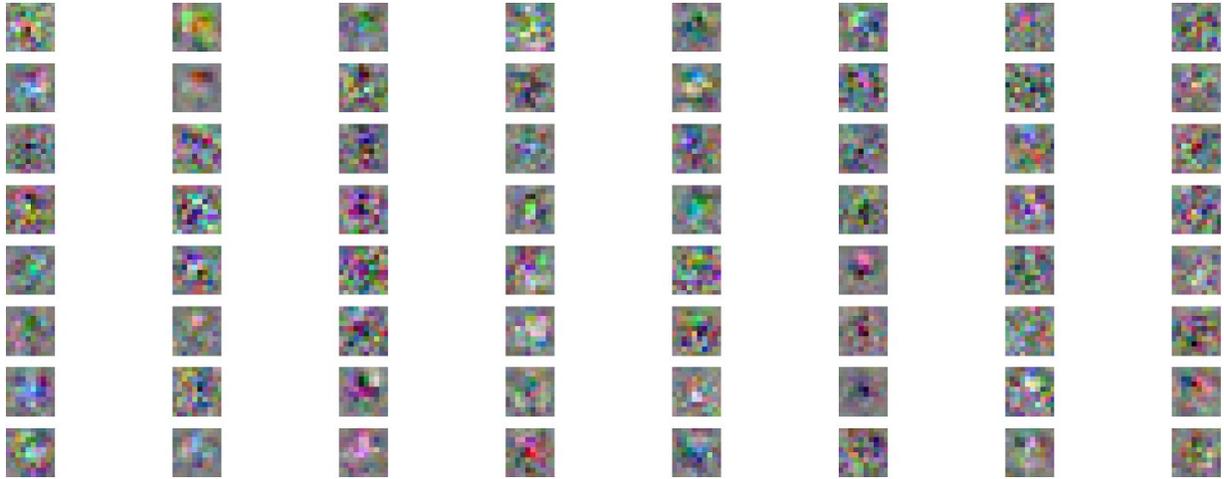

Iteration 400,000, SPC: $W^{sp}$ with 64 kernels of size 10x10

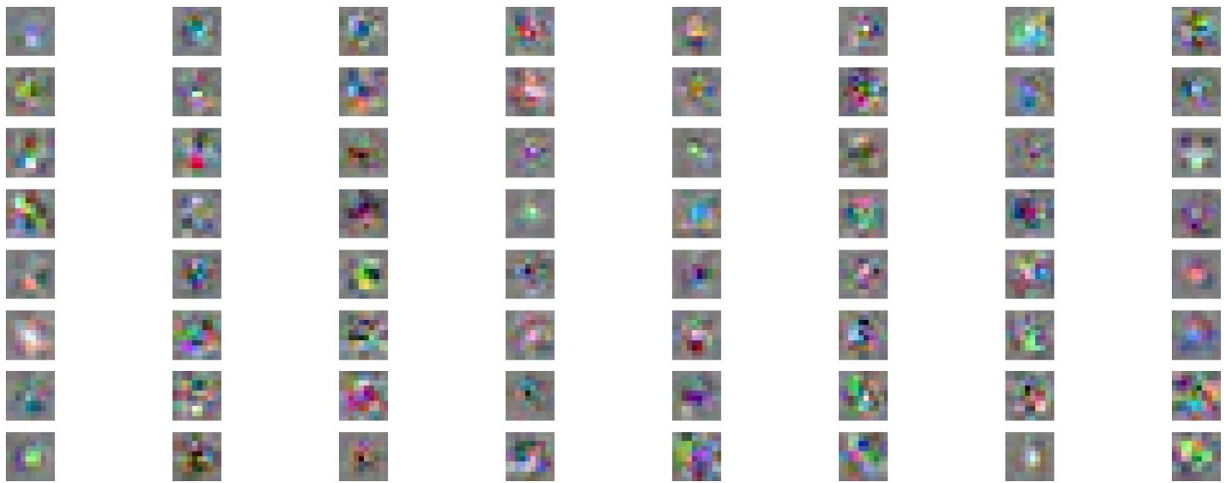

Iteration 400,000, SPC initialized to convolution NN resize: $W^{sp}$ with 64 kernels of size 10x10

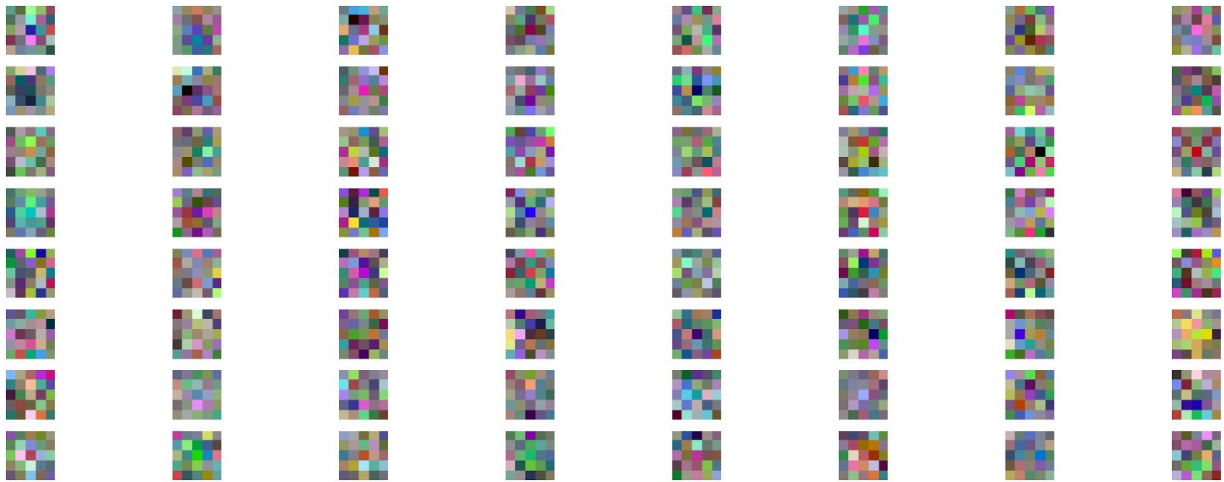

Iteration 400,000, resize convolution: $W^{sp}$ with 64 kernels of size 5x5

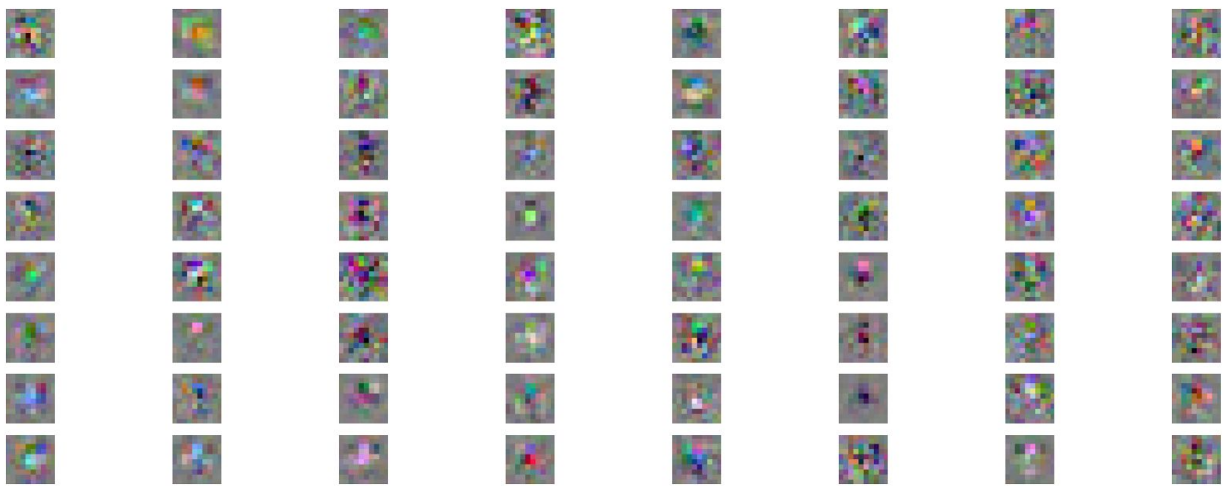

Iteration 1,000,000, SPC: $W^{sp}$ with 64 kernels of size 10x10

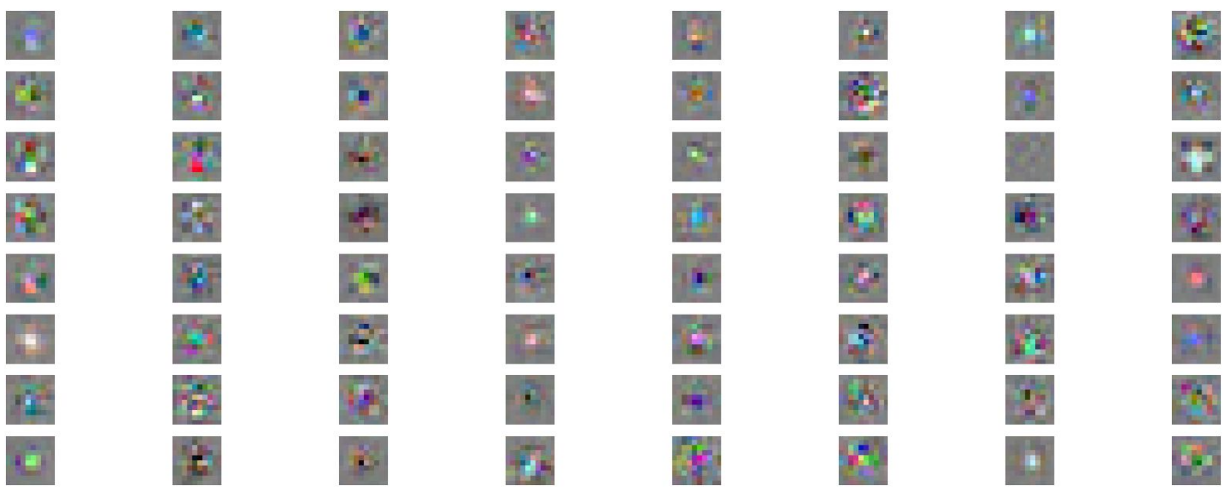

Iteration 1,000,000, SPC initialized to convolution NN resize: $W^{sp}$ with 64 kernels of size 10x10

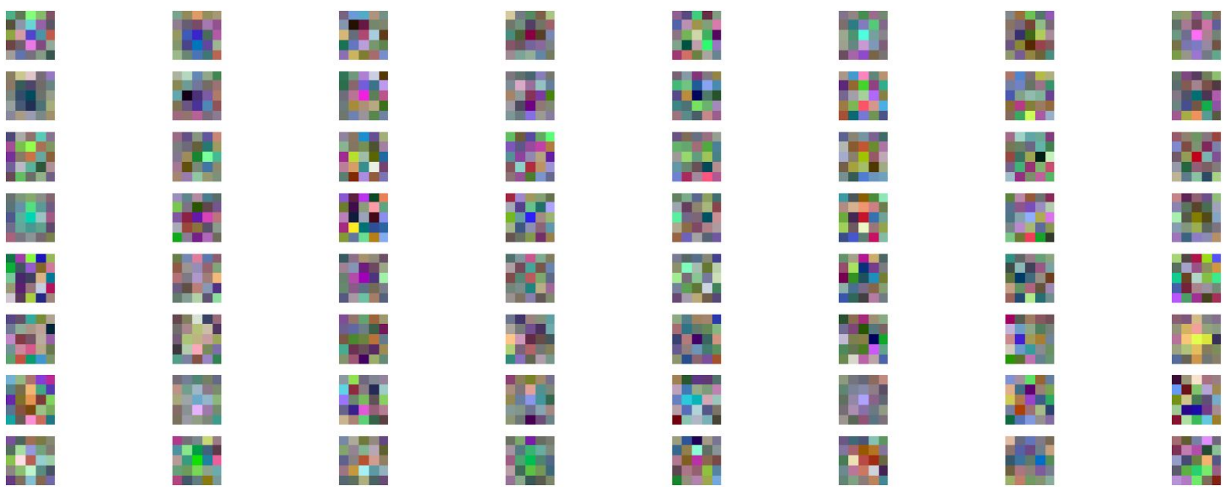

Iteration 1,000,000, resize convolution: $W^{sp}$ with 64 kernels of size 5x5

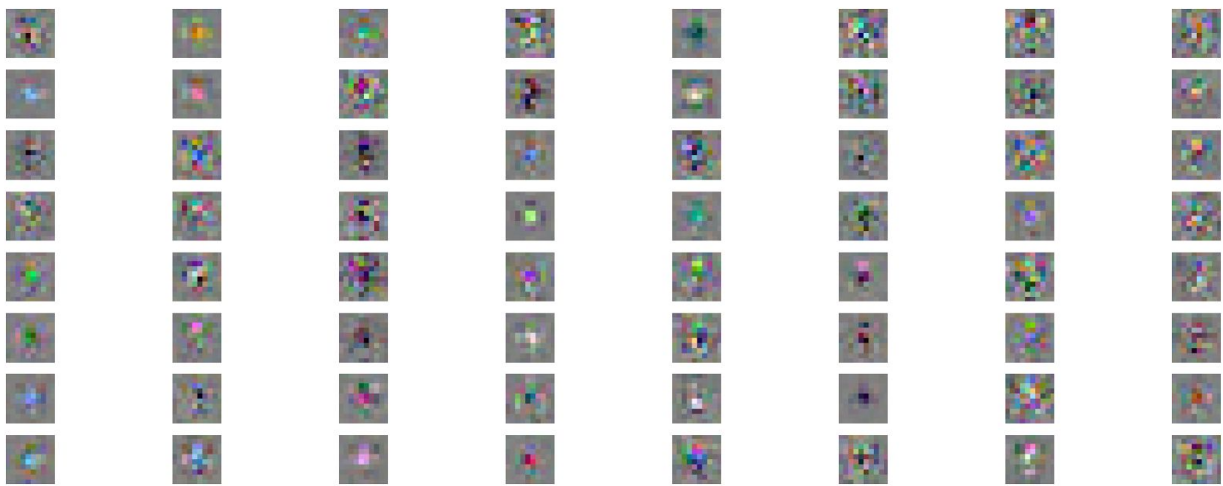

Iteration 2,000,000, SPC: $W^{sp}$ with 64 kernels of size 10x10

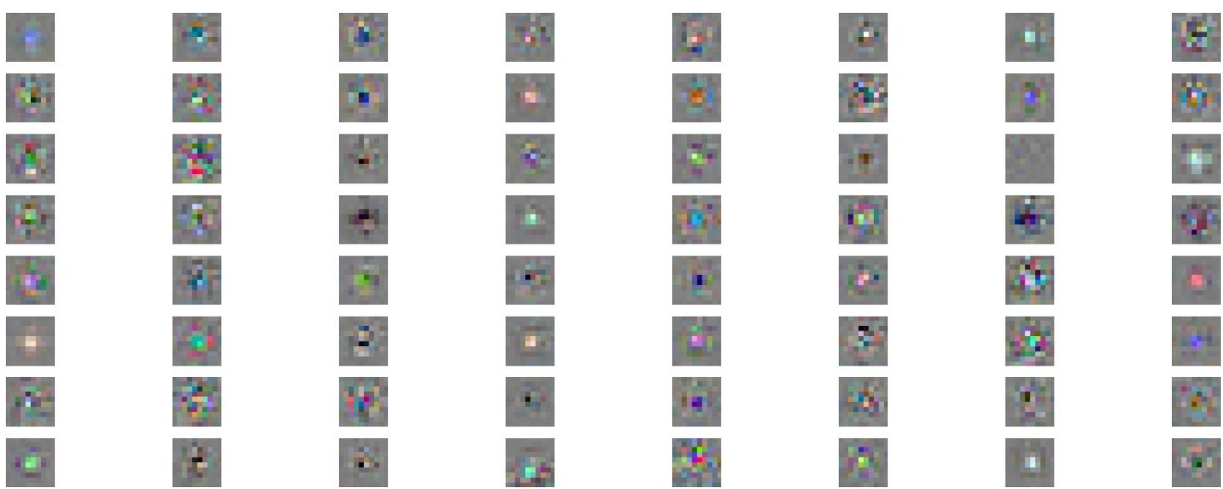

Iteration 2,000,000, SPC initialized to convolution NN resize: $W^{sp}$ with 64 kernels of size 10x10

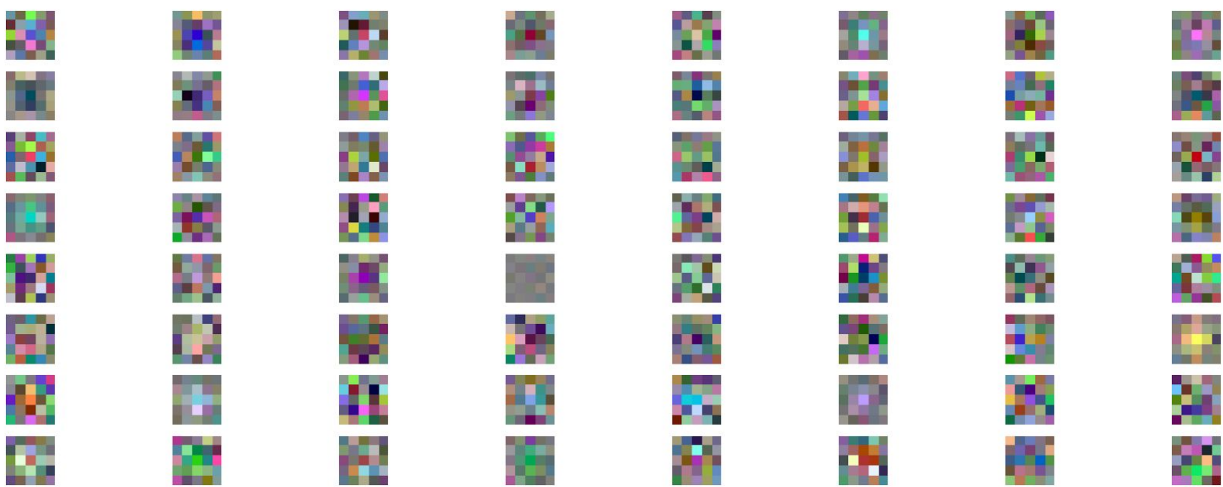

Iteration 2,000,000, resize convolution: $W^{sp}$ with 64 kernels of size 5x5

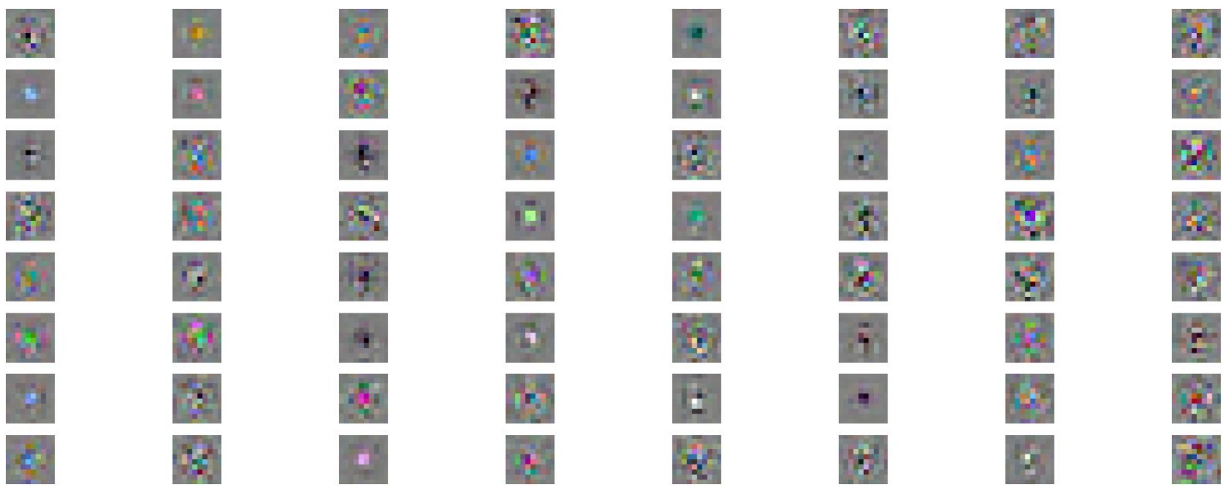

Iteration 4,000,000, SPC: $W^{sp}$ with 64 kernels of size 10x10

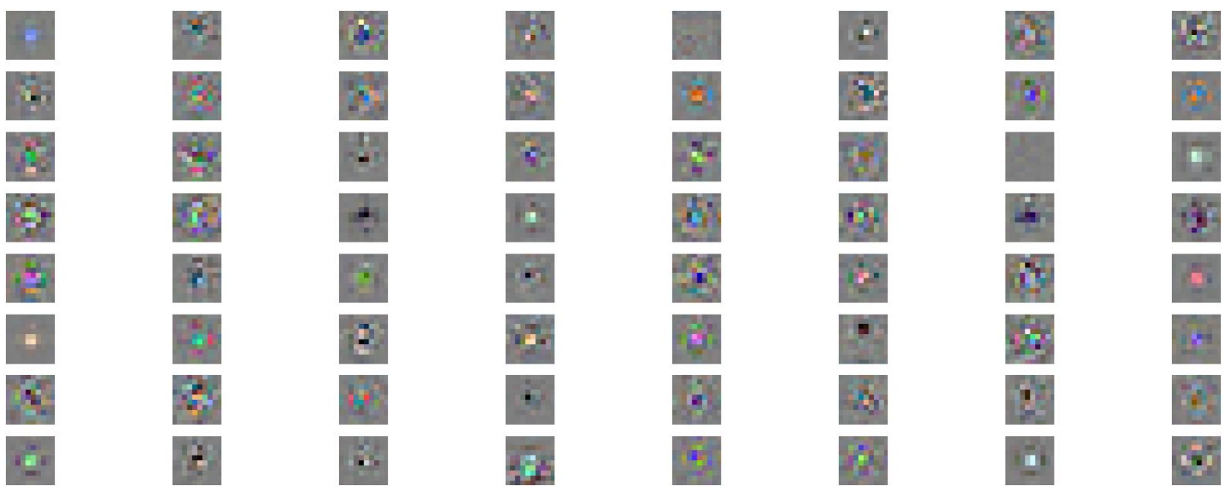

Iteration 4,000,000, SPC initialized to convolution NN resize: $W^{sp}$ with 64 kernels of size 10x10

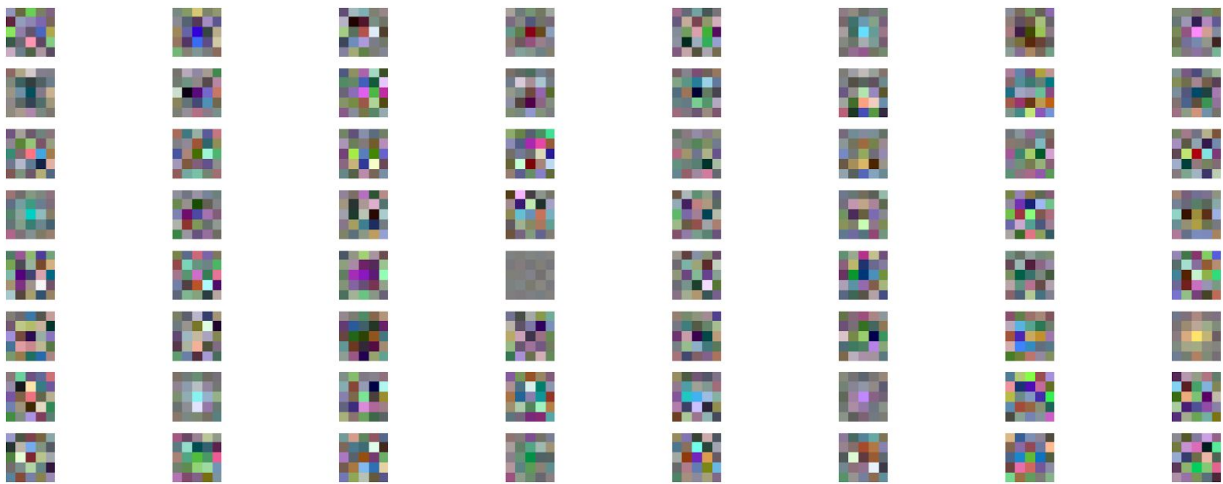

Iteration 4,000,000, resize convolution: $W^{sp}$ with 64 kernels of size 5x5

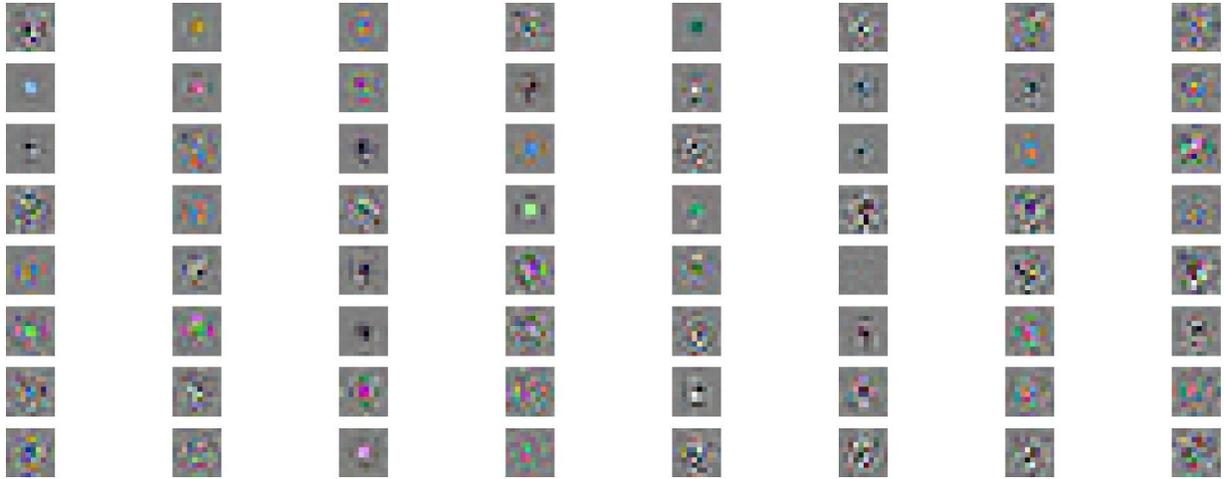

Iteration 8,000,000, SPC: $W^{sp}$ with 64 kernels of size 10x10

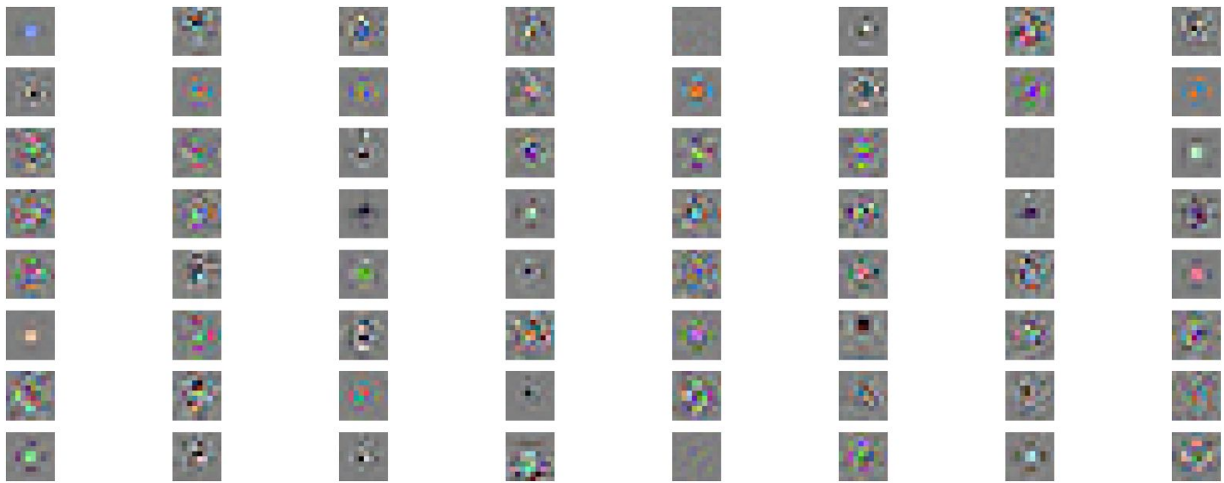

Iteration 8,000,000, SPC initialized to convolution NN resize: $W^{sp}$ with 64 kernels of size 10x10

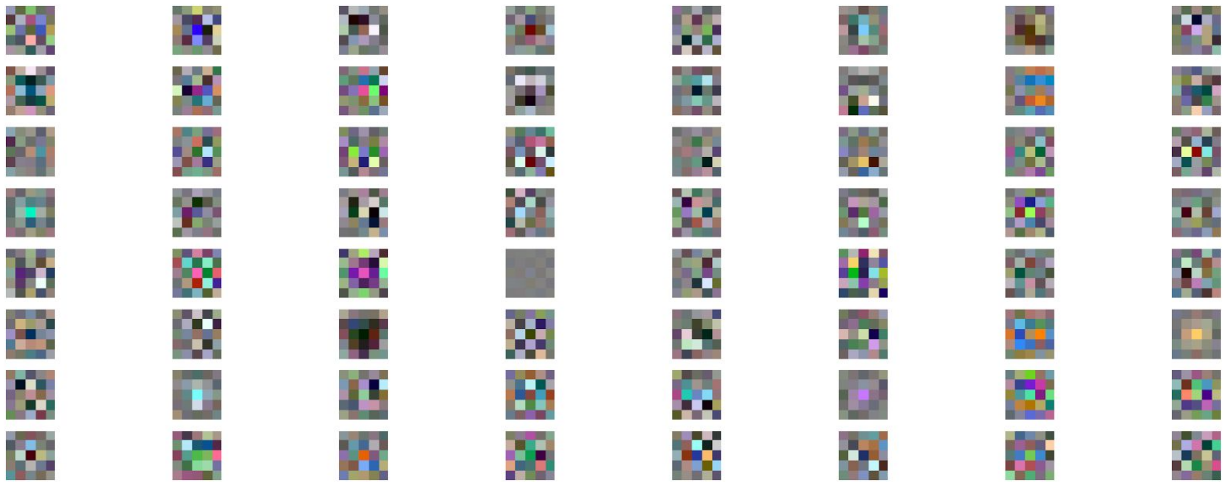

Iteration 8,000,000, resize convolution: $W^{sp}$ with 64 kernels of size 5x5